\documentclass[11pt]{article}
\usepackage[utf8]{inputenc}

\usepackage[preprint]{acl}

\usepackage{times}
\usepackage[T1]{fontenc}        
\usepackage[utf8]{inputenc}     
\usepackage{microtype}          
\usepackage{inconsolata}        

\usepackage{latexsym}
\usepackage{xurl} 
\usepackage[centertags]{amsmath}

\usepackage{booktabs}
\usepackage{multirow}

\usepackage{graphicx}
\usepackage{subcaption}
\usepackage{float}

\usepackage{tcolorbox}
\tcbuselibrary{breakable,skins}
\usepackage{fancyvrb}
\usepackage{fvextra}            

\usepackage{afterpage}
\title{H$^2$R: Hierarchical Hint-Based Reward for Mathematical 
Reasoning via Reinforcement Learning}

\author{
  \textbf{He Tu\textsuperscript{1,*}},
  \textbf{Siye Li\textsuperscript{1,*}},
  \textbf{Hongyi Wen\textsuperscript{1}},
  \\
  \\
  \textsuperscript{1}New York University Shanghai
  \\
  \small{\textsuperscript{*}Equal contribution.}
  \\
  \small{
    \textbf{Emails:}
    \href{mailto:HELEN_EMAIL}{ht2604@nyu.edu},
    \href{mailto:SIYE_EMAIL}{sl10962@nyu.edu},
    \href{mailto:hongyi.wen@nyu.edu}{hongyi.wen@nyu.edu}
  }
}

\begin{document}
\maketitle
 
\afterpage{%
\begin{figure*}[t]
    \centering
    \includegraphics[page=2, width=\textwidth,
        trim={0 4cm 0 4cm}, clip]{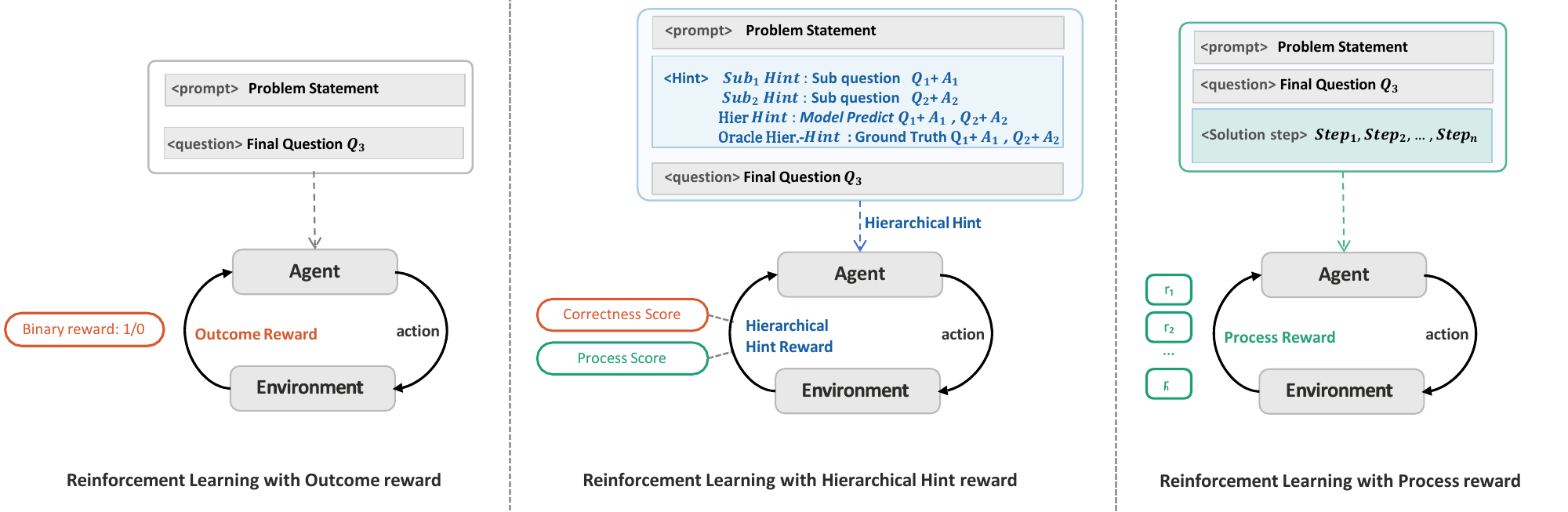}
    \caption{Comparison of three reward signal paradigms on the
    same mathematical reasoning problem. Given an identical model
    output, ORM assigns a single binary reward based solely on
    final-answer correctness, providing no signal on intermediate
    reasoning quality. PRM assigns dense per-step scores requiring
    costly step-level annotation. \textbf{H$^2$R achieves a middle
    ground}: a hint derived from sub-questions is provided during
    training to steer the model toward rigorous reasoning, and a
    judge model evaluates both answer correctness and process
    quality, yielding a hybrid reward without any step-level
    annotation.}
    \label{fig:reward_comparison}
\end{figure*}
}

\begin{abstract}
Reinforcement Learning with Verifiable Rewards has shown great promise in enhancing mathematical reasoning in large language models. However, existing approaches provide limited intermediate supervision for complex, multi-step reasoning tasks. To bridge this gap, we propose \textbf{Hierarchical Hint-based Reward}, a framework that leverages the intrinsic sub-question structure of mathematical problems as structured intermediate conditioning signals during RLVR training. Unlike chain-of-thought prompting, these hierarchical hints serve as logically grounded intermediate milestones that effectively reduce the reasoning horizon toward the final answer. To validate this framework, we introduce \textbf{HiMath}, a benchmark of hierarchically structured mathematical problems. Across diverse LLMs, hierarchical hints consistently improve reasoning performance even without task-specific training. When integrated into RL training via H$^2$R, they boost Qwen2.5-Math-7B accuracy on HiMath from 30.4\% to 39.1\% using hierarchical hints, and up to 45.6\% with Oracle Hierarchical Hints. This pronounced improvement demonstrates that incorporating reliable intermediate checkpoints into reward design is essential for sustaining complex mathematical reasoning.
\end{abstract}

\section{Introduction}
 
Large Language Models (LLMs) have demonstrated remarkable
potential in complex reasoning tasks, such as mathematical
problem-solving~\citep{cobbe2021gsm8k, hendrycksmath2021}, code
generation~\citep{jain2024livecodebench, jimenez2024swebench}, and
logical inference~\citep{srivastava2023beyond}. Among these,
mathematical reasoning serves as a critical frontier: it requires
sustained, multi-step logical trajectories where every
intermediate decision critically impacts the final validity. While
current models perform well on standard benchmarks, reliably
solving hierarchically structured, high-level mathematical
problems remains a significant challenge for existing
architectures~\citep{hendrycksmath2021, gsmsymbolic}.
 
Reinforcement Learning with Verifiable Rewards (RLVR) has become
the primary paradigm for bridging this gap. The prevailing
approach relies on sparse rewards, where a positive
signal is assigned only if the final answer is
correct~\citep{deepseekr1, wen2025rlvr}. While scalable, sparse
rewards suffer from a fundamental credit assignment
problem: a model may receive zero reward for a nearly correct
derivation that fails only at the final step, while another model
is positively reinforced for reward hacking, arriving at the
correct answer through flawed or degenerate
reasoning~\citep{skalse2025rewardhacking}. To mitigate this,
dense rewards (e.g., Process Reward
Models~\citep{lightman2023lets, luo2024omegaprm,
zheng2025surveyprocessrewardmodels}) provide feedback at each
step. However, dense rewards face a severe scalability bottleneck:
obtaining step-level annotations requires prohibitive human labor
or expensive teacher-model synthesis, and PRMs do not consistently
generalize across
domains~\citep{chen2025mathematicalreasoningcodegeneralization}.
This creates a structural gap: sparse rewards are too coarse, while
dense rewards are too costly, and both largely ignore the
hierarchical decomposition encoded in many existing problems.

In this paper, we identify an overlooked structural property of
complex mathematics: multi-part sub-question
structure naturally serves as a principled resolution to this dilemma.
Competitive exams typically decompose a problem into logically
ordered sub-questions ($Q_1, Q_2, Q_3$), where earlier results
serve as prerequisite milestones for later ones. We argue that
these sub-questions are \textbf{hierarchical hints}: naturally
occurring, verifiable intermediate goals that encode the intended
reasoning trajectory without requiring any additional manual
step-level labeling. Building on this observation, we propose
\textbf{Hierarchical Hint-based Reward (H$^2$R)}, a
reward-shaping framework that integrates sub-question correctness
as intermediate reward signals within the
GRPO~\citep{deepseek-math} training objective. Importantly,
H$^2$R operates at the level of reward design, decomposing the
long-horizon reasoning trajectory into a sequence of verifiable
sub-goals, requiring no step-level annotation or architectural
change. Unlike inference-time decomposition methods, the hierarchical hints
are used only during training and are removed entirely at test time;
the trained policy therefore receives neither hints nor decomposition
prompts at deployment. As illustrated in Figure~\ref{fig:reward_comparison},
conditioning the model on a sub-question hint lifts the reward
from $0.0$ to $0.8$ by anchoring its reasoning to a critical
intermediate result.

H$^2$R bridges sparse outcome rewards and dense process supervision by
leveraging the problem's logical structure to guide the model toward
semantically meaningful intermediate states. This hierarchy provides
a natural basis for intermediate supervision, while a lightweight
LLM judge further enriches the binary outcome reward with process-level
assessment~\citep{zheng2023judging, li2024llmsasjudges}. To instantiate
and evaluate H$^2$R, we introduce \textbf{HiMath}, a curated dataset of
hierarchically structured mathematical problems.

Our experiments 
reveal three main findings: (1) in a purely inference-only 
setting, hierarchical hints consistently improve accuracy across 
eight diverse LLMs, confirming that their value is intrinsic to 
logical structure rather than task-specific fine-tuning; (2) when 
used as training-time supervision signals in RLVR, hierarchical 
hints drive consistent accuracy gains over standard GRPO, with 
all improvements persisting at test time \textit{without any hint 
dependency};(3) H$^2$R generalizes consistently to MATH500 and AIME across 
three model families, outperforming GRPO, DAPO, and SAGE baselines 
by substantial margins,showing that the gains are not confined to the HiMath training distribution.

Our contributions are as follows:
\begin{itemize}
    \item We introduce \textbf{HiMath}, a contamination-resistant 
    benchmark of hierarchically structured mathematical problems, 
    each decomposing into three logically chained sub-questions. 
  
    \item We demonstrate that sub-question-derived hints 
    consistently improve reasoning accuracy across diverse LLMs 
    even without task-specific training, establishing that their 
    value is intrinsic to logical structure.

    \item We propose \textbf{H$^2$R}, incorporating a hybrid 
    reward design that combines answer correctness with 
    process-level evaluation. Gains persist at test time 
    \textit{without any hint dependency}.
\end{itemize}
 
\section{Related Work}
 
\paragraph{Reinforcement Learning for LLM Reasoning.}
Prompt-based approaches such as Chain-of-Thought
prompting~\citep{wei2023cot},
Self-Consistency~\citep{wang2023selfconsistency}, and
Tree-of-Thought~\citep{yao2023tree} leverage prompts to encourage
step-by-step reasoning. Building on these foundations,
learning-based methods have advanced reasoning through supervised
fine-tuning on high-quality reasoning
datasets~\citep{yu2023outcomesupervised, yu2025longshort,
yang2024qwen2}.
 
Recently, RLVR demonstrated that ``thinking'' behaviors can be
induced through verifiable
rewards~\citep{deepseekr1, wang2025reinforcement}. Methods such as
GRPO~\citep{shao2024deepseekmath} and PPO~\citep{schulman2017ppo}
have shown that models can self-discover reasoning paths when
provided with binary reward signals. Research directions have
shifted toward addressing reward
hacking~\citep{skalse2025rewardhacking}, extending reasoning to
multimodal settings~\citep{zhang2025r1vl}, and improving
efficiency~\citep{chae2025selfplay, guo2025g2rpoa}. Our work
builds on this RLVR paradigm, addressing the sparsity of
outcome-only rewards by introducing structured intermediate
conditioning.
 
\paragraph{Reward Signal Design: Sparse vs.\ Dense.}
 
ORMs and PRMs represent two ends of a fundamental spectrum. ORM
supervises reasoning solely based on final-answer
correctness~\citep{cobbe2021gsm8k}. While R1-style models
demonstrate that ORM alone can induce self-correction
behaviors~\citep{deepseekr1, openr1}, early reasoning errors
remain invisible until the end of
trajectories~\citep{sutton1998rl}. PRMs address this by providing
dense, step-level
feedback~\citep{lightman2023lets, khalifa2025prm, yao2026prl,
li2025prmqvalue}, but require costly step-by-step annotations or Monte Carlo
estimation~\citep{luo2024omegaprm, wang2024mathshepherd,
xiong2025minimalist}, and do not consistently outperform ORM on
strong RL-trained models~\citep{feng2025prmnecessary}.
Recent hybrid approaches seek a middle
ground~\citep{jia2025hybrid, setlur2025rewarding, yu2025reasoning}.
H$^2$R is positioned in this spirit: rather than training a full
PRM or relying on binary rewards alone, we leverage the intrinsic
sub-question hierarchy of complex mathematical problems as
verifiable intermediate checkpoints, augmented by an LLM judge for
process quality assessment.
 \begin{figure*}[t]
    \centering
    \includegraphics[page=3, width=\textwidth,
        trim={0 3.3cm 0 4cm}, clip]{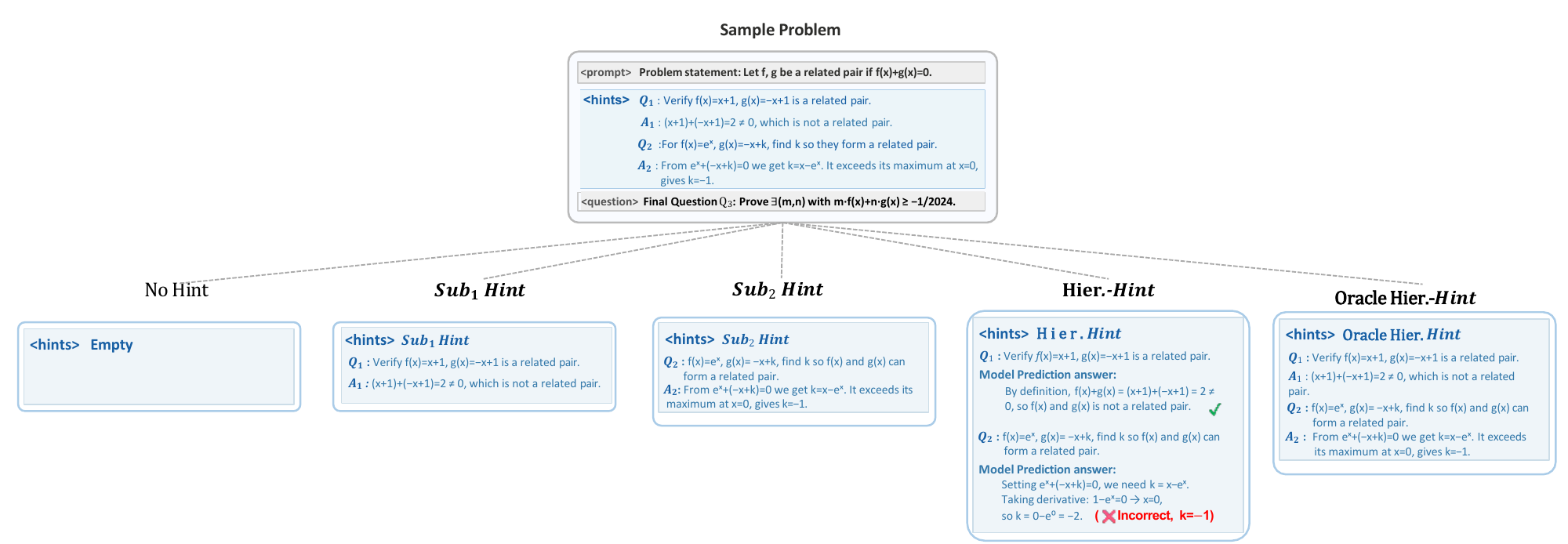}
    \caption {Overview of the H$^2$R framework. Five hint conditions are
constructed by varying the \texttt{<hints>} field: No-Hint provides
no intermediate context; Sub$_1$-Hint and Sub$_2$-Hint provide the
ground-truth solution to $Q_1$ or $Q_2$, respectively; Hier.-Hint
provides model-generated solutions to both sub-questions; and Oracle
Hier.-Hint provides both ground-truth solutions as an oracle upper
bound. At each iteration, the policy receives a hint-conditioned
prompt and H$^2$R reward, and GRPO updates the policy toward
high-reward trajectories.}
    \label{fig:H2R_framework}
\end{figure*}

\section{Hierarchical Hint Framework}
 \subsection{The HiMath Dataset}
\label{sec:dataset}

We introduce \textbf{HiMath}, a structured, multi-part mathematics dataset specifically engineered to evaluate hierarchical reasoning. Inspired by the formatting of competitive examinations, each problem introduces a novel mathematical concept directly within its problem statement. 

Each HiMath problem contains three sequentially dependent sub-questions ($Q_1$, $Q_2$, and $Q_3$) of increasing difficulty. This hierarchy originates from the original problem design, with human reference solutions providing verifiable intermediate supervision for H$^2$R. HiMath comprises 151 problems across functions, sequences, sets, and geometry. Rather than maximizing scale, it serves as a focused testbed for studying hierarchical reasoning. Its novel-concept format encourages reasoning from first principles and reduces reliance on memorization~\citep{wu2025reasoningmemorizationunreliableresults}. Data collection details are provided in Appendix~\ref{app:dataset}.


Although HiMath is relatively compact, its scale is intentional:
like targeted benchmarks such as BIG-Bench Hard~\citep{srivastava2023beyond}
and AIME~\citep{hendrycksmath2021}, it emphasizes focused problem
selection over sheer volume. HiMath further prioritizes structural depth
and contamination resistance: its explicit sub-question structure
provides natural, reference-anchored intermediate supervision for H$^2$R,
while its novel-concept design reduces reliance on memorization
~\citep{wu2025reasoningmemorizationunreliableresults}. Despite being
trained on only 105 HiMath problems, the resulting models also improve
on the held-out MATH500 and AIME benchmarks
(Table~\ref{tab:generalization}), indicating that the gains are not
confined to the HiMath training distribution.

\paragraph{Difficulty Analysis}
\label{sec:difficulty}

To demonstrate that HiMath problems present a genuine intellectual challenge rather than tasks solvable via pattern matching or memorization, we evaluate a broad selection of state-of-the-art LLMs on $Q_3$ without providing any hint context. 
The empirical results paint a clear picture: even the most capable closed-source reasoning models fall well short of ceiling performance. Specifically, o3-mini~\citep{openai2025o3o4mini} achieves only $66.9\%$, Gemini-3.1-pro~\citep{google2025gemini31} scores $63.5\%$, and Gemini-2.5-pro~\citep{comanici2025gemini} reaches $59.3\%$. 
The performance gap widens substantially for open-source models. Qwen2.5-Math-7B-Instruct and Qwen2.5-7B-Instruct~\citep{qwen2.5} both achieve around $20.7\%$, Qwen2.5-Math-7B drops to $17.9\%$, and Qwen3-4B reaches only $12.4\%$~\citep{yang2025qwen3}. By comparison, the standard MATH500 pass rates for these identical open-source models regularly exceed $75\%$.

These findings highlight two properties that make HiMath suitable for
evaluating H$^2$R: its difficulty leaves substantial headroom for
intermediate supervision to improve performance, while its novel-concept
design reduces reliance on memorization~\citep{wu2025reasoningmemorizationunreliableresults}.
Moreover, models trained on the 105 HiMath training problems also improve
on held-out MATH500 and AIME benchmarks
(Table~\ref{tab:generalization}), suggesting that the gains are not confined to the HiMath training distribution.

 
\begin{figure*}[t]
    \centering
    \includegraphics[width=\textwidth]{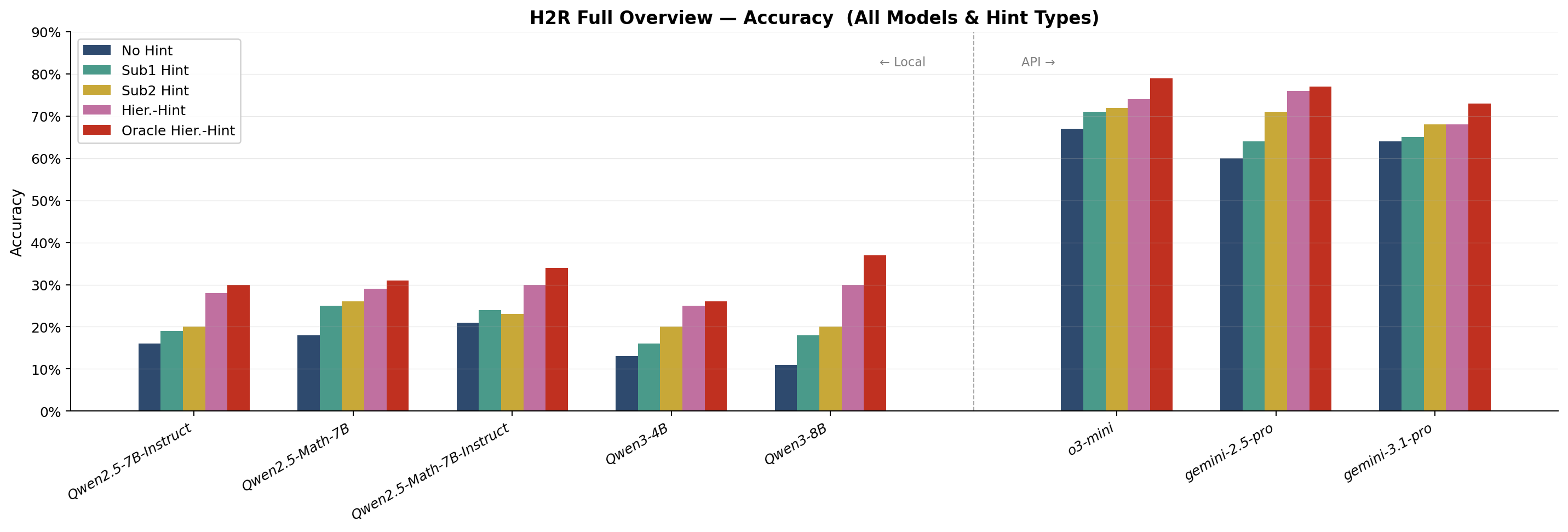}
    \caption{Inference-only accuracy of eight untrained checkpoints under five hint conditions, with hints given at inference time and no RL training. Hints consistently improve performance across model scales and architectures. Unlike these results, H$^2$R-trained policies receive no hints at inference. Differences between adjacent Qwen model scales are not statistically significant (Appendix~\ref{app:cross-model-significance}).}

    \label{fig:baseline}
\end{figure*}

\subsection{Inference with Hierarchical Hint}

We design a hierarchy of five hint conditions, illustrated in
Figure~\ref{fig:H2R_framework}, to investigate how structured
intermediate conditioning affects the model's ability to solve $Q_3$. These conditions span a spectrum ranging from a complete absence of context, denoted as No-Hint, to full oracle supervision via oracle Hier.-Hint, with three intermediate levels designed to isolate distinct reasoning factors. Specifically, Sub$_1$-Hint and Sub$_2$-Hint provide ground-truth solutions to individual sub-questions, establishing whether foundational or proximal intermediate results are more load-bearing for $Q_3$. Meanwhile, Hier.-Hint supplies solutions generated by a stronger external model for both sub-questions simultaneously, reflecting a realistic deployment scenario that operates without oracle supervision. Comparing the Hier.-Hint against oracle Hier.-Hint reveals whether performance differences stem from genuine reasoning transfer or merely from the quality of the intermediate context provided. By varying hint informativeness in a controlled manner, we directly attribute performance shifts to specific reasoning failure modes.

We evaluate hint effectiveness in an inference-only
setting. Figure~\ref{fig:baseline} presents results across
eight models under five hint conditions. Accuracy gains range from
$+2.8\%$ to $+13.8\%$ for Sub$_1$-Hint and $+0.7\%$ to $+17.4\%$
for Sub$_2$-Hint, highlighting the value of the logical structure
encoded in hierarchical hints. Full results are in
Appendix~\ref{app:inference-only-full}. Pairwise bootstrap comparisons
across the Qwen model family reveal no statistically significant
differences between adjacent model scales
(Appendix~\ref{app:cross-model-significance}).


\subsection{Analysis of Hint Effectiveness}
\label{sec:hint_effectiveness}
 
\paragraph{Logical Proximity Drives Hint Effectiveness.}
Sub$_1$-Hint provides early-stage contextual information useful
but not immediately actionable for $Q_3$. Sub$_2$-Hint occupies a
more pivotal position: it establishes the key intermediate
constraint that $Q_3$ directly builds upon. This structural
difference explains why the marginal accuracy gain from
Sub$_2$-Hint is nearly $50\%$ larger than Sub$_1$-Hint on average
($+8.0\%$ vs.\ $+5.5\%$ across eight models).

\paragraph{The Role of Hier.-Hint.}
Across all models, Hier.-Hint consistently outperforms both Sub$_1$-Hint and 
Sub$_2$-Hint, even though those two conditions provide oracle answers to individual 
sub-questions. This finding suggests that the performance advantage is primarily driven 
by the completeness of the intermediate context. Simultaneously providing solutions 
to both $Q_1$ and $Q_2$ offers a more comprehensive reasoning scaffold than either 
alone, and this holds true even when solutions are imperfect.

Hier.-Hint underperforms oracle Hier.-Hint across all models, highlighting the value of accurate intermediate solutions. Errors in $Q_1$ or $Q_2$ can propagate to $Q_3$ (Appendix~\ref{app:error-propagation}).

\subsection{RL with Hierarchical Hint Reward}
 
The consistent inference-only gains in Figure~\ref{fig:baseline} motivate the use of hierarchical hints as structured signals during RL training. Unlike sparse outcome-only rewards, hint-conditioned training mitigates the credit assignment problem by providing verifiable intermediate milestones aligned with the problem's logical structure. We therefore incorporate hints of varying informativeness into the GRPO~\citep{shao2024deepseekmath} framework, guiding the policy toward more reliable reasoning trajectories.
 
\paragraph{Reward Design.}
H$^2$R employs a two-component reward function that jointly
evaluates answer correctness and reasoning quality. For the
$i$-th model response, the reward $s_i$ is computed as:
 
\begin{equation}
\label{eq:reward}
s_i = \left\{\,
\begin{array}{ll}
\alpha + (1-\alpha) \cdot \min\!\left(1,\dfrac{p_i}{\tau}\right)
    & \text{if } c_i = 1 \\[10pt]
\vphantom{\dfrac{p_i}{\tau}}0
    & \text{if } c_i = 0
\end{array}
\right.
\end{equation}
 
\noindent where $c_i \in \{0,1\}$ denotes final-answer
correctness, $p_i \in [1,5]$ is the reasoning process quality
score assessed by an LLM judge~\citep{zheng2023judging},
$\alpha \in (0,1)$ is the base reward weight for answer
correctness, and $\tau$ is a normalization threshold that maps
the discrete process score to a continuous bonus range. We set
$\alpha = 0.6$ and $\tau = 4$ throughout, yielding a reward in
$[0.6, 1.0]$ for correct responses whose process quality ranges
from adequate to excellent.

\paragraph{Hard Gate and Reward Hacking Prevention.}
We enforce a hard gate $s_i = 0$ whenever $c_i = 0$, regardless of reasoning quality. Without gating, a model can receive non-zero reward for plausible-sounding but mathematically invalid reasoning, since the process score $p_i$ evaluates structural fluency rather than absolute logical validity. As $Q_3$ is typically stated as an explicit claim to be proved, a model can mimic correct reasoning by copying the conclusion and padding it with invalid steps. A naive hybrid reward would therefore incentivize such pattern matching over genuine mathematical reasoning~\citep{skalse2025rewardhacking}.

Table~\ref{tab:gate-ablation} validates this design. Removing the hard gate degrades performance: Soft Gate ($s_i > 0$ permitted when $c_i = 0$) reaches $41.30\%$ ($-4.35$ pp), while No Gate (pure linear combination) falls to $39.13\%$ ($-6.52$ pp). Without gating, $34.2\%$ of incorrect responses receive high process scores ($p_i \geq 3$), creating false training signals that the hard gate eliminates. Both non-gated variants fall in the suboptimal region of the hyperparameter grid (Appendix~\ref{app:hard-gate}), supporting the hard gate as a principled component of the reward design.

Note that the inference-only results in Figure~\ref{fig:baseline} differ from those of the trained policy: an untrained Qwen2.5-7B-Instruct checkpoint given oracle hints at inference reaches only about $30\%$, whereas the H$^2$R-trained policy reaches $41.30\%$ with no hints at evaluation, showing that the improvement is retained in the learned policy.

\begin{table*}[t]
\centering
\footnotesize
\begin{tabular}{llcccc}
\toprule
\textbf{Model} & \textbf{Method} & \textbf{Training Hint} &
\textbf{HiMath} & \textbf{MATH500} & \textbf{AIME24 / 25} \\
 & & & (avg@3) & (pass@1) & (avg@5) \\
\midrule

\multirow{8}{*}{Llama-3.2-3B-Instruct}
 & Initial checkpoint   & {---}              & 11.11          & 44.7           & 6.5 / 0.6           \\
 & GRPO       & No-Hint            & 11.67          & 54.6           & 10.0 / 0.7          \\
 & DAPO       & No-Hint            & 13.04          & 57.0           & 6.7 / 0.5           \\
 & SFT        & Oracle Hier.-Hint  & 13.04          & 53.8           & 9.5 / ---           \\
 & SAGE-Light & Self-Hint          & 8.69           & 54.1$^\dagger$ & 8.8 / 0.9           \\
 & SAGE       & Self-Hint          & 13.04          & 56.3$^\dagger$ & 9.2 / 0.8           \\
 & H$^2$R     & Hier.-Hint         & 19.57          & 59.5           & 12.1 / 1.2          \\
 & H$^2$R     & Oracle Hier.-Hint  & \textbf{21.73} & \textbf{61.2}  & \textbf{13.3 / 1.7} \\
\midrule

\multirow{8}{*}{Qwen2.5-7B-Instruct}
 & Initial checkpoint   & {---}              & 15.86          & 75.7           & 13.8 / 6.7          \\
 & GRPO       & No-Hint            & 19.57          & 77.8           & 15.0 / 10.67        \\
 & DAPO       & No-Hint            & 21.74          & 82.0           & 16.31 / 12.00       \\
 & SFT        & Oracle Hier.-Hint  & 32.61          & 77.1           & 14.8 / ---          \\
 & SAGE-Light & Self-Hint          & 19.56          & 79.9$^\dagger$ & 17.1 / 11.7         \\
 & SAGE       & Self-Hint          & 23.91          & 80.0$^\dagger$ & 16.0 / 12.5         \\
 & H$^2$R     & Hier.-Hint         & 34.78          & 84.5           & 22.00 / 15.33       \\
 & H$^2$R     & Oracle Hier.-Hint  & \textbf{41.30} & \textbf{84.8}  & \textbf{23.90 / 17.30} \\
\midrule

\multirow{8}{*}{Qwen3-4B-Instruct}
 & Initial checkpoint   & {---}              & 25.16          & 90.6           & 22.1 / 13.40        \\
 & GRPO       & No-Hint            & 32.88          & 96.0           & 24.20 / 14.40       \\
 & DAPO       & No-Hint            & 34.28          & 93.4           & 29.33 / 12.00       \\
 & SFT        & Oracle Hier.-Hint  & 39.13          & 95.6           & 23.8 / ---          \\
 & SAGE-Light & Self-Hint          & 26.08          & 95.1$^\dagger$ & 27.6 / 14.7         \\
 & SAGE       & Self-Hint          & 28.26          & 95.4$^\dagger$ & 25.1 / 14.9         \\
 & H$^2$R     & Hier.-Hint         & 47.82          & 97.06          & 34.30 / 17.33       \\
 & H$^2$R     & Oracle Hier.-Hint  & \textbf{52.17} & \textbf{97.6}  & \textbf{36.00 / 18.87} \\
\bottomrule
\end{tabular}

\caption{Performance on HiMath and public benchmarks across three model
families, where all models are evaluated without hints at test time.
GRPO, DAPO, and SFT are controlled baselines trained on the same 105
HiMath problems, while SAGE and SAGE-Light are included as external
reference points trained under a different data regime. The MATH500
results marked with $^\dagger$ are sourced from
\citet{liao2026selfhintinglanguagemodelsenhance}, while all other
results come from our own evaluation. For the hint conditions, Hier.-Hint
uses hints generated by o3-mini, whereas Oracle Hier.-Hint relies on
expert-annotated ground-truth solutions to establish an upper bound only.
The metrics are reported as the mean accuracy over three independent
training seeds for HiMath avg@3, a single greedy decoding for MATH500
pass@1, and the mean over five samples at temperature $0.7$ for AIME avg@5.}

\label{tab:generalization}
\end{table*}
\begin{table}[t]
\begin{center}
\small
\begin{tabular}{lcccc}
\toprule
\textbf{Gate} & \textbf{Acc.} & \textbf{Avg.P} &
\textbf{Incorr.\ $p_i{\geq}3$ (\%)} & \textbf{Grad.} \\
\midrule
No Gate   & 39.13 & 3.21 & 34.2 & None \\
Soft Gate & 41.30 & 3.45 & 18.7 & Partial \\
Hard Gate & \textbf{45.65} & \textbf{3.89} & \textbf{0.0} & Full \\
\bottomrule
\end{tabular}
\caption{Gate design ablation on Qwen2.5-Math-7B with Oracle Hier.-Hint. ``Incorr.\ $p_i \geq 3$'' measures the fraction of incorrect responses receiving a high process score and directly captures the reward-hacking rate. The hard gate eliminates this false training signal while achieving the highest accuracy.}
\label{tab:gate-ablation}
\end{center}
\end{table}

\section{Experiments}
\label{sec:hint-RL}

\paragraph{Evaluation of Proof-Style Questions.}
Since $Q_3$ is typically phrased as a claim to be proved rather than a value to be computed, final-answer correctness $c_i$ is assessed by the LLM judge rather than exact string matching. The judge is instructed to verify whether the submitted proof is logically complete and valid. Each HiMath problem is accompanied by a complete human-written reference solution, which serves as an anchor for the evaluation. The judge assesses whether a submitted proof is consistent with the reference in both its derivation and final conclusion, rather than evaluating it against an open-ended notion of correctness.

To assess evaluator sensitivity, we additionally evaluated the same responses with GPT-4o and DeepSeek-R1 as independent judges. Although the judges differ slightly in absolute accuracy, they preserve the same ordering across No-Hint, Hier.-Hint, and Oracle Hier.-Hint, supporting the robustness of the observed treatment effect. We further conducted a blinded human audit on a subset of the evaluated responses, including additional cases where the automated judges disagreed. Detailed cross-judge and human-validation results are reported in Appendix~\ref{app:judge-validation}.

\subsection{Models and Training Configuration}
 
We evaluate H$^2$R on HiMath using Qwen2.5-Math-7B and
Qwen2.5-7B-Instruct for the hint-condition ablation. For
out-of-domain generalization experiments, we follow the model
selection of \citet{liao2026selfhintinglanguagemodelsenhance} and
use: Llama-3.2-3B-Instruct~\citep{meta2024llama3},
Qwen2.5-7B-Instruct~\citep{yang2024qwen2}, and
Qwen3-4B-Instruct~\citep{yang2025qwen3}, representing low,
moderate, and high levels of math-focused optimization. All models
are trained with GRPO~\citep{shao2024deepseekmath} within the
\texttt{verl}~\citep{sheng2024hybridflow} framework.  Detailed hyperparameters are provided in Appendix~\ref{app:hyperparams}.

\subsection{Main Results}

Full hint condition ablation results including Avg.\ Reward and Avg.\ Process Score are reported in Appendix~\ref{app:hint-ablation}. In summary, accuracy increases monotonically from No-Hint to Oracle Hier.-Hint for both models (Table~\ref{tab:hint-training}), confirming that structured intermediate conditioning during RL effectively guides the model toward better reasoning paths without any hint dependency at test time.

The primary practical result is Hier.-Hint: for Qwen2.5-Math-7B, Hier.-Hint achieves $39.13\%$ accuracy ($+8.70$ pp over baseline), using only LLM-generated intermediate solutions with no oracle supervision. Oracle Hier.-Hint, which requires expert-annotated sub-question solutions, serves as an \textit{oracle upper bound} at $45.65\%$ ($+15.22$ pp), establishing the performance ceiling available when intermediate supervision is perfect. Qwen2.5-7B-Instruct shows a similar pattern, improving from $19.57\%$ to $34.78\%$ under Hier.-Hint and $41.30\%$ under Oracle Hier.-Hint. Across model families, training seeds, and benchmarks, H$^2$R consistently improves performance over No-Hint training.

To assess the statistical reliability of these gains, we conducted paired cluster bootstrap resampling over the 46 test problems with 10,000 replicates. Compared with No-Hint, Hier.-Hint improves accuracy by $15.22$ percentage points (95\% CI: $[4.3, 26.1]$, $p=0.006$), while Oracle Hier.-Hint improves accuracy by $21.74$ percentage points (95\% CI: $[9.4, 33.4]$, $p<0.001$), confirming that both gains are statistically significant.

Table~\ref{tab:generalization} evaluates three model families on public
benchmarks. H$^2$R consistently outperforms the controlled GRPO, DAPO,
and SFT baselines under the same evaluation setting. We also compare
H$^2$R with SAGE~\cite{liao2026selfhintinglanguagemodelsenhance}, which
uses approximately 220K training problems, compared with just 105
HiMath problems for H$^2$R. On HiMath, H$^2$R (Hier.-Hint) surpasses
SAGE by $6.53$, $10.87$, and $19.56$ percentage points across the
three models, without oracle supervision during training. Additional
comparisons with Random-Hint and Unordered-Hint isolate the contribution
of hierarchical reward design under the same HiMath training setting.

Compared with SFT trained on the same 105 HiMath problems using oracle hints, H$^2$R performs better across all evaluated models and benchmarks, despite using generated hints in its practical setting. Its stronger gains on held-out benchmarks suggest that reward-based training better transfers the benefits of hierarchical supervision beyond HiMath.

\subsection{Random-Hint and Unordered-Hint Ablations}

To distinguish the effect of meaningful hierarchical structure from the effect of additional supervision or context, we introduce two training-time controls. Random-Hint pairs each problem with another problem's gold hints, preserving the hint format, labels, and supervision volume while removing semantic relevance. Unordered-Hint uses the problem's own gold hints but reverses their order and replaces hierarchical labels with a generic marker, preserving the hint content while removing its hierarchical organization. All hints are used only during training and are removed at inference.

\begin{table*}[t]
\centering
\small
\setlength{\tabcolsep}{8pt}
\begin{tabular}{llcccc}
\toprule
\textbf{Model} & \textbf{Benchmark} & \textbf{No-Hint} &
\textbf{Random} & \textbf{Unordered} & \textbf{Hier.} \\
\midrule

\multirow{3}{*}{Llama-3.2-3B-Instruct}
& HiMath  & $11.67$ & $10.87$ & $15.22$ & $19.57$ \\
& MATH500 & $54.6$  & $54.8$  & $57.4$  & $59.5$ \\
& AIME24  & $10.0$  & $10.1$  & $11.0$  & $12.1$ \\

\midrule

\multirow{3}{*}{Qwen2.5-7B-Instruct}
& HiMath  & $19.57$ & $21.74$ & $26.09$ & $34.78$ \\
& MATH500 & $77.8$  & $78.2$  & $80.8$  & $84.5$ \\
& AIME24  & $15.0$  & $15.3$  & $17.7$  & $22.0$ \\

\midrule

\multirow{3}{*}{Qwen3-4B-Instruct}
& HiMath  & $32.88$ & $32.61$ & $41.30$ & $47.82$ \\
& MATH500 & $96.0$  & $96.1$  & $96.4$  & $97.1$ \\
& AIME24  & $24.20$ & $24.50$ & $27.6$  & $34.30$ \\

\bottomrule
\end{tabular}

\caption{Ablation of hint relevance and hierarchical organization across
three model families and three benchmarks. Hints are used only during
training and removed at inference.}
\label{tab:hint-structure-ablation}
\end{table*}

The results form an ablation ladder. Across all nine settings, Random-Hint provides no consistent improvement over the No-Hint baseline, differing by at most one problem in HiMath and $0.4$ points in MATH500. Since Random-Hint preserves the same reward function, gate, and supervision volume as Hier.-Hint, its lack of improvement indicates that the gains do not arise from additional training signal or hint format. Unordered-Hint consistently falls between Random-Hint and Hier.-Hint, recovering roughly half of the gain. Because it preserves the original hint content while removing its ordering and hierarchical labels, the gap between Unordered-Hint and Hier.-Hint isolates the contribution of hierarchical organization, while the gap between Random-Hint and Unordered-Hint reflects the importance of semantic relevance. The same ordering is observed on the held-out MATH500 and AIME24 benchmarks, indicating that the benefit is not limited to the HiMath training distribution.

\subsection{Empirical Validation of the Hybrid Reward Design}

We do not include a direct baseline with a standard process reward model (PRM) because its conventional construction is not directly applicable to HiMath's proof-style questions. In particular, HiMath $Q_3$ problems do not provide a canonical answer string from which intermediate proof states can be automatically labeled, so adapting a PRM would require additional judge-based annotation of intermediate steps and a substantially different supervision pipeline. A detailed analysis of this constructability issue and the associated annotation cost is provided in Appendix~\ref{app:prm-constructability}.

The concurrent trajectory of Avg.\ Process Score in
Table~\ref{tab:hint-training} demonstrates that our reward
function successfully steers optimization without triggering
reward hacking. For Qwen2.5-Math-7B, the average process score
climbs from $2.81$ under No-Hint to $3.89$ under Oracle Hier.-Hint. This
growth confirms that H$^2$R penalizes
superficial pattern-matching and incentivizes rigorous,
verifiable intermediate milestones. The gate design ablation in
Appendix~\ref{app:hard-gate} validates this choice.

\section{Discussion} \paragraph{Intermediate-state Reliability.} The Hier.-Hint condition mirrors how human students solve multi-step problems by conditioning on their self-generated intermediate results, thereby exposing realistic failure modes such as error propagation and limited self-correction. Our error-propagation analysis (Appendix~\ref{app:error-propagation}) suggests that the primary bottleneck is not the model's ability to use intermediate context once provided, but its ability to \textit{generate} reliable intermediate states in the first place. This finding motivates hint generation itself as a trainable objective and moves H$^2$R toward a fully self-supervised framework. \paragraph{Generalization to Policy Optimization.} More broadly, H$^2$R requires only that problems decompose into verifiable sub-goals, a structure that arises naturally across STEM domains such as physics derivations, proof-based geometry, and multi-step programming. We conjecture that expert-encoded problem hierarchies could provide intermediate training signals in these settings, though validation beyond mathematics remains future work. This work focuses on reward formulation rather than optimization: by fixing the GRPO framework, we isolate reward structure as the independent variable and attribute improvements to hierarchical hint-based reward design. Developing algorithms tailored to multi-stage reasoning, such as curriculum scheduling over hint conditions or hint-aware policy optimization, remains an important direction.

\section{Conclusion}

We presented H$^2$R, a reward-shaping framework that uses hierarchical sub-question structures to provide intermediate supervision for LLM reasoning. By combining final-answer correctness with LLM-judged process quality, H$^2$R yields a practical hybrid reward signal that can improve reasoning without requiring oracle supervision at test time. Through Hier.-Hint training, we show consistent gains across three model families, three random seeds, and three benchmarks. Our analysis further suggests that the key bottleneck in multi-step reasoning is not whether models can use intermediate context when it is available, but whether they can generate reliable intermediate states in the first place. Poor intermediate hints induce error propagation and substantially limit downstream performance. To support future work on this problem, we introduce HiMath, a benchmark with explicit sub-question hierarchies. Together, H$^2$R and HiMath provide a foundation for studying more fine-grained process rewards and training algorithms tailored to robust multi-stage reasoning.

\section*{Limitations}
\paragraph{Dataset Scale.}
HiMath prioritizes reasoning depth and contamination resistance over scale: each problem introduces a novel mathematical concept and chains three logically dependent sub-questions, ensuring substantial structural richness per example. H$^2$R trained on HiMath improves performance on the held-out MATH500 and AIME24 benchmarks, indicating that the observed gains are not confined to the HiMath training distribution. However, these results do not establish broad or general reasoning ability, and evaluation on a wider range of tasks and datasets is needed to assess the scope of the learned behavior. Broader empirical coverage across more diverse problem types therefore remains an important direction for future work.

\paragraph{Automated Evaluation.}
Our evaluation relies on automated grading, combining final-answer matching with LLM-as-judge assessment of reasoning quality. To assess evaluator robustness, we evaluated the same responses with o3-mini, GPT-4o, and DeepSeek-R1. Although the judges differ slightly in absolute accuracy, they preserve the same ordering across No-Hint, Hier.-Hint, and Oracle Hier.-Hint, with the corresponding treatment effects remaining statistically significant. We further conducted a blinded human audit of 25 responses, including 20 randomly sampled responses and 5 cases selected from judge disagreements. The two authors jointly judged 19 of the 20 randomly sampled responses as correct or incorrect in agreement with the automated evaluation, while 23 of the 25 responses were consistent in the pooled audit. Nevertheless, automated evaluation may not capture all nuances of mathematical reasoning, and broader human validation would provide an additional quality check.

\paragraph{PRM Baseline.}
We do not include a direct process reward model (PRM) baseline because conventional PRM construction is not directly applicable to HiMath's proof-style questions. In particular, HiMath Q3 problems do not provide a canonical answer string from which intermediate proof states can be automatically labeled. Adapting an existing PRM pipeline would therefore require additional judge-based annotation of intermediate proof steps and a substantially different supervision pipeline. This constructability issue, together with the associated annotation and computational cost, makes a faithful in-domain PRM comparison beyond the scope of the current study. Developing proof-oriented PRMs for this setting remains an important direction for future work.

\section*{Ethics Statement}
While our work enhances the hierarchical reasoning capabilities
of LLMs in complex mathematics, we recognize that such
advancements are dual-use. We are committed to responsible
development of AI. All pre-trained models used in this work 
are publicly available under open-source licenses. Our code 
and the HiMath dataset will be released for academic use 
upon paper acceptance.
\bibliography{custom}
 
\appendix

\section{HiMath Dataset}
\label{app:dataset}
 
\paragraph{The New Definition Question.}
HiMath is built around \textit{New Definition} questions, a
problem format originating from competition mathematics. Such questions introduce a novel mathematical concept within the exam itself, requiring models to reason from first principles rather than recall a memorized procedure. Their primary purpose is to assess mathematical abstraction, logical reasoning, and exploratory problem-solving, making them well-suited for probing the reasoning limits of LLMs.
 
A typical problem consists of a formal definition followed by
three subquestions of increasing difficulty. Earlier subquestions
establish results logically prerequisite to solving the final
one, instantiating the hierarchical hint structure central to
H$^2$R.
 
\paragraph{Design Principles.}
HiMath is constructed according to four principles:
 
\begin{itemize}
    \item \textbf{Difficulty and Reasoning Depth.} New Definition
    questions require multi-step inference across overlapping
    domains, providing deeper insight than standard benchmarks.
 
    \item \textbf{Novelty and Contamination Resistance.} Each
    problem introduces a previously unseen concept, requiring
    models to reason from first principles rather than recall
    memorized patterns.
 
    \item \textbf{Hierarchical Sub-question Structure.} The
    three-part design progresses from simple to complex, with
    earlier sub-questions establishing foundational results
    essential to the final one.
 
    \item \textbf{Annotation Quality and Diversity.} Each problem
    includes an official ground-truth solution with detailed
    annotations. The dataset spans functions, sequences, sets,
    and geometry.
\end{itemize}
 
\paragraph{Data Collection.}
We collect New Definition questions from public official
examination repositories and established educational websites. PDF materials were converted to
\LaTeX{} using Mathpix-based OCR followed by manual verification.
 
\paragraph{Example New Definition Question}
\label{app:example_problem}
 
\begin{tcolorbox}[
  title={Example: New Definition Question from HiMath},
  colback=white,
  colframe=black,
  fonttitle=\bfseries,
  breakable
]
\medskip
\textbf{Problem Statement.}\\
Let $A = \{1,2,3,4,5,6,7,8\}$ and define
$M = \{(x,y) \mid x \in A,\, y \in A\}.$
Select $n$ points from $M$ to form a sequence
$(x_1,y_1), (x_2,y_2), \ldots, (x_n,y_n).$
Two consecutive points must satisfy either
$|x_{i+1}-x_i| = 3,\; |y_{i+1}-y_i| = 4$
or
$|x_{i+1}-x_i| = 4,\; |y_{i+1}-y_i| = 3.$
Such a sequence is called a \emph{k-sequence}.
 
\begin{enumerate}
  \item If the first term is $(3,3)$, determine the second term.
  \item Determine whether $(3,2)$ and $(4,4)$ can both appear
  in a $k$-sequence $\tau$ where odd-indexed terms have
  $x_i \in \{1,2,7,8\}$ and even-indexed terms have
  $x_i \in \{3,4,5,6\}$.
  \item Prove that it is impossible for all elements in $M$ to
  form a $k$-sequence.
\end{enumerate}
\end{tcolorbox}

\section{Experimental Setting}
\label{app:hyperparams}
We split the 151 problems into training and test sets at an 
approximately 7:3 ratio, yielding 105 training problems and 
a held-out test set of 46 problems. The split is performed 
at the problem level to ensure no cross-split contamination: 
all three subquestions of a given 
problem are assigned exclusively to one split. All hint 
conditions are evaluated on this identical test set 
throughout all experiments.

Following the experimental design of 
\citet{liao2026selfhintinglanguagemodelsenhance}, we select 
backbone LLMs spanning varying degrees of mathematical 
specialization. Llama-3.2-3B-Instruct~\citep{meta2024llama3} 
represents a low level of math-focused optimization, while 
Qwen2.5-7B-Instruct~\citep{yang2024qwen2} offers moderate 
general and mathematical reasoning capabilities and is also 
used for our detailed hint-condition ablation studies. 
Qwen2.5-Math-7B is a math-specialized model used for 
ablation extensions to verify domain-specific persistence. Qwen3-4B-Instruct-2507
(\texttt{Qwen/Qwen3-4B-Instruct-2507})~\citep{yang2025qwen3} represents a high 
level of mathematical proficiency, having been extensively 
aligned via reinforcement learning. We use the updated Qwen3-4B-Instruct-2507 checkpoint rather than
the original Qwen3-4B checkpoint; throughout the paper, we refer
to this model as Qwen3-4B-Instruct for brevity.All models are optimized using the Group Relative Policy 
Optimization (GRPO) algorithm~\citep{shao2024deepseekmath} 
implemented within the \texttt{verl} 
framework~\citep{sheng2024hybridflow}, with \texttt{o3-mini} 
serving as the primary judge for our hybrid reward function. We apply a maximum response length of 2,048 tokens; 
responses exceeding this limit are filtered out. This cutoff 
affects only a negligible fraction of generations, is 
uniformly distributed across all hint conditions, and 
introduces no systematic bias. For optimization, we follow 
standard GRPO practices with a group size of $G=2$, a clip 
epsilon of $0.2$, and a learning rate of $1\times10^{-6}$ 
with an AdamW optimizer under cosine learning rate 
scheduling. Each training run takes approximately 3--4 hours 
of wall-clock time on 4 NVIDIA H100 GPUs; the 
cumulative computational cost across all models, hint 
conditions, and 3 independent random seeds amounts to 
approximately 600 GPU hours. Full implementation scripts, 
dataset splits, and reproducibility instructions will be 
publicly released via GitHub upon acceptance.

\section{Training and Judge Prompt Templates}
\label{app:train_prompt}

\begin{tcolorbox}[
  title={Training Prompt Template},
  colback=white, colframe=black, fonttitle=\bfseries,
  breakable, 
  top=3pt, bottom=3pt,
  width=\columnwidth,          
  boxsep=2pt, left=4pt, right=4pt 
]
\begin{Verbatim}[
  baselinestretch=0.85, 
  fontsize=\footnotesize,     
  breaklines=true
]
<|im_start|>system
You are a helpful and rigorous mathematics assistant. Reason step by step. Clearly state the final answer using \boxed{} inside <answer> </answer> tags.
<|im_end|>
<|im_start|>user
<prompt>{problem}</prompt>
<hints>{hint_block}</hints>
<question>{subquestion3}</question>
<|im_end|>
<|im_start|>assistant
<think>
\end{Verbatim}
\end{tcolorbox}
\begin{tcolorbox}[
  title={Judge Prompt Template},
  colback=white, colframe=black, fonttitle=\bfseries,
  breakable, 
  top=3pt, bottom=3pt,
  width=\columnwidth,          
  boxsep=2pt, left=4pt, right=4pt
]
\begin{Verbatim}[
  baselinestretch=0.85, 
  fontsize=\footnotesize,      
  breaklines=true
]
Evaluate along TWO dimensions:
1. correctness: 0 or 1
2. process_score: 1-5
Rule: If correctness=0, process_score <= 3.
Return ONLY JSON:
{"correctness": 0|1,
 "process_score": int,
 "comment": "..."}
\end{Verbatim}
\end{tcolorbox}

\section{Main Experimental Results and Statistical Analysis}
\subsection{Statistical Significance.}
We assess the statistical significance of the accuracy gains using paired cluster bootstrap resampling over the 46 test problems. Each bootstrap replicate samples test problems with replacement and retains all model predictions associated with each sampled problem, preserving the pairing between the compared conditions. We use 10,000 bootstrap replicates to estimate the distribution of the accuracy difference between each hint condition and No-Hint. For Hier.-Hint, the estimated improvement is $15.22$ percentage points with a 95\% confidence interval of $[4.3, 26.1]$ percentage points ($p=0.006$). For Oracle Hier.-Hint, the estimated improvement is $21.74$ percentage points with a 95\% confidence interval of $[9.4, 33.4]$ percentage points ($p<0.001$).

\subsection{Per-Seed Results}
\label{app:per-seed}

Table~\ref{tab:per-seed} reports per-seed accuracy on the HiMath
test set (46 problems) for all hint conditions.
 
\begin{center}
\small
\begin{tabular}{llcccc}
\toprule
\textbf{Model} & \textbf{Cond.} & \textbf{S1} &
\textbf{S2} & \textbf{S3} & \textbf{Mean} \\
\midrule
\multirow{5}{*}{\shortstack[l]{Qwen2.5\\-Math-7B}}
 & No-Hint & 23.91 & 32.61 & 34.78 & 30.43 \\
 & Sub$_1$ & 28.26 & 30.43 & 39.13 & 32.61 \\
 & Sub$_2$ & 30.43 & 39.13 & 41.30 & 36.96 \\
 & Hier     & 32.61 & 41.30 & 43.48 & 39.13 \\
 & Oracle    & 43.48 & 45.65 & 47.83 & 45.65 \\
\midrule
\multirow{5}{*}{\shortstack[l]{Qwen2.5\\-7B-Inst.}}
 & No-Hint & 13.04 & 21.74 & 23.91 & 19.57 \\
 & Sub$_1$ & 19.57 & 28.26 & 30.43 & 26.09 \\
 & Sub$_2$ & 21.74 & 30.43 & 32.61 & 28.26 \\
 & Hier     & 28.26 & 36.96 & 39.13 & 34.78 \\
 & Oracle    & 34.78 & 43.48 & 45.65 & 41.30 \\
\bottomrule
\end{tabular}
\captionof{table}{Per-seed accuracy (\%) on HiMath test set (46 problems); max deviation 2 problems ($4.35\%$).}
\label{tab:per-seed}
\end{center}
\vspace{2pt}

\subsection{Cross-Model Statistical Significance}
\label{app:cross-model-significance}

To assess whether the observed ordering across model scales is statistically meaningful, we applied the same paired cluster bootstrap procedure to pairwise comparisons between adjacent Qwen model variants on the HiMath test set. Comparing Qwen2.5-7B-Instruct with Qwen3-4B-Instruct yields a difference of $3.31$ percentage points (95\% CI: $[-4.0, 10.6]$, $p=0.365$), and comparing Qwen3-4B-Instruct with Qwen3-8B-Instruct yields a difference of $1.32$ percentage points (95\% CI: $[-6.0, 8.6]$, $p=0.727$). Neither difference is statistically significant, indicating that performance on HiMath is comparable across these adjacent scales. The hint-induced gains are therefore consistent across model families and are not driven by model scale.

\subsection{Hint Condition Ablation Results}
\label{app:hint-ablation}
\begin{table*}[t]
\centering
\small
\begin{tabular}{llcccc}
\toprule
\textbf{Model} & \textbf{Training Hint} & \textbf{Test Hint} &
\textbf{Accuracy (\%)} & \textbf{Avg.\ Reward} &
\textbf{Avg.\ Process Score} \\
\midrule
\multirow{5}{2cm}{Qwen2.5-Math-7B}
 & No-Hint (GRPO Baseline) & None & 30.43 & 0.22 & 2.81 \\
 & $+$ Sub$_1$-Hint        & None & 32.61 & 0.23 & 2.96 \\
 & $+$ Sub$_2$-Hint        & None & 36.96 & 0.28 & 3.25 \\
 & $+$ Hier.-Hint           & None & 39.13 & 0.31 & 3.48 \\
 & $+$ Oracle Hier.-Hint  & None & \textbf{45.65} & \textbf{0.35} & \textbf{3.89} \\
\midrule
\multirow{5}{2cm}{Qwen2.5-7B-Instruct}
 & No-Hint (GRPO Baseline) & None & 19.57 & 0.18 & 2.43 \\
 & $+$ Sub$_1$-Hint        & None & 26.09 & 0.21 & 2.61 \\
 & $+$ Sub$_2$-Hint        & None & 28.26 & 0.26 & 2.79 \\
 & $+$ Hier.-Hint           & None & 34.78 & 0.29 & 3.12 \\
 & $+$ Oracle Hier.-Hint  & None & \textbf{41.30} & \textbf{0.33} & \textbf{3.23} \\
\bottomrule
\end{tabular}
\caption{Performance of models trained under different hint
conditions, \textbf{all evaluated without hints at test time}.
Accuracy: avg@3 over 3 independent random seeds; std $\leq$ 2
problems  across all conditions. Oracle Hier.-Hint serves as an
oracle upper bound using expert-curated sub-question solutions.
Avg.\ Reward and Avg.\ Process Score are averaged over all test
samples.}
\label{tab:hint-training}
\end{table*}

Table~\ref{tab:hint-training} reports the full hint-condition 
ablation on HiMath for Qwen2.5-Math-7B and 
Qwen2.5-7B-Instruct, evaluated without hints at test time.

\section{Reward Design Analysis}

\subsection{Reward Instantiation.}
We set $\alpha = 0.6$ and $\tau = 4$, yielding:
\begin{equation}
s_i = \begin{cases}
0.6 + 0.4 \cdot \min\!\left(1, \dfrac{p_i}{4}\right)
    & \text{if } c_i = 1 \\[6pt]
0 & \text{if } c_i = 0
\end{cases}
\end{equation}

\subsection{Hard Gate Ablation}
\label{app:hard-gate}
We further analyze the role of the correctness gate through Soft Gate and No Gate variants. On Qwen2.5-Math-7B with Oracle Hier.-Hint, Soft Gate reduces accuracy from $45.65\%$ to $41.30\%$, while No Gate reduces it to $39.13\%$. The corresponding reward-hacking rates, measured as the fraction of incorrect responses with $p_i \geq 3$, are $18.7\%$ and $34.2\%$, respectively, compared with $0.0\%$ under the hard gate. These results show that allowing process reward on incorrect trajectories introduces substantial false training signals. The full hyperparameter sensitivity analysis further shows that both non-gated variants remain in the suboptimal region of the search grid, supporting the hard gate as a principled safeguard rather than a tuned choice.

\subsection{Reward Hyperparameter Ablation.}
Table~\ref{tab:hyperparam-ablation} reports sensitivity analysis
over the $\{\alpha, \tau\}$ grid under the Oracle Hier.-Hint setting using Qwen2.5-Math-7B evaluated with avg@3.
\begin{center}
\small
\begin{tabular}{lccc}
\toprule
 & $\tau = 3$ & $\tau = 4$ & $\tau = 5$ \\
\midrule
$\alpha = 0.5$ & 40.58 & 42.75 & 42.03 \\
$\alpha = 0.6$ & 42.75 & \textbf{45.65} & 42.75 \\
$\alpha = 0.7$ & 40.58 & 42.03 & 39.37 \\
\bottomrule
\end{tabular}
\captionof{table}{Accuracy (\%) across the $\{\alpha,\tau\}$ grid
(Qwen2.5-Math-7B, Oracle Hier.-Hint, avg@3).}
\label{tab:hyperparam-ablation}
\end{center}
\vspace{2pt}
 
The optimal $\alpha = 0.6$, $\tau = 4$ balances correctness
dominance with meaningful process gradient.

\subsection{PRM Constructability}
\label{app:prm-constructability}
Standard process reward models (PRMs) are not directly applicable to
HiMath because the dataset does not provide canonical intermediate
states or step-level labels for its proof-style questions. In
particular, the $Q_3$ subquestions require open-ended mathematical
proofs rather than a fixed answer string from which intermediate
reasoning states can be automatically segmented and labeled.

Adapting a conventional PRM pipeline would therefore require an
additional judge-based annotation process to evaluate intermediate
proof states. Such an adaptation would substantially change the
supervision pipeline used in our experiments. For example, a
proof-oriented adaptation of existing PRM methods would require
constructing partial-proof annotations and repeatedly evaluating
intermediate states against the reference solution, introducing a
substantial additional annotation and inference cost.

This constructability issue is particularly important for HiMath
because our existing judge evaluates a complete response using both
its final correctness and its consistency with the human-written
reference solution. Converting this evaluation into dense
step-level supervision would require additional judgments for
intermediate proof prefixes rather than simply evaluating the final
response. We therefore leave a direct in-domain PRM comparison to
future work and instead validate the dense component of our hybrid
reward through the process-score trajectory and the hard-gate
ablation reported in the main text.

\section{LLM Judge Validation}
\label{app:judge-validation}

\paragraph{Reference-Anchored Evaluation.}
Following \citet{zheng2023judging}, we use o3-mini as the primary judge in a zero-shot setting. For each response, the judge outputs $(c_i,p_i,e_i)$, where $c_i \in \{0,1\}$ denotes final-answer correctness, $p_i \in [1,5]$ is the process score, and $e_i$ is a brief justification. We enforce $p_i \leq \tau-1=3$ whenever $c_i=0$ to prevent incorrect responses from receiving a high process reward.

The evaluation is reference-anchored rather than open-ended. Every HiMath problem is accompanied by a complete human reference solution. The judge evaluates whether a submitted response is consistent with the reference in both its derivation and final conclusion, rather than judging correctness against an unconstrained notion of mathematical validity. As a basic validation of the correctness component, when provided with ground-truth solutions, the judge assigns $p_i \geq 4$ in $91.3\%$ of cases.

\paragraph{Cross-Judge Validation.}
To assess potential self-preference from o3-mini's dual role as both Hier.-Hint generator and reward judge, we re-scored the same 138 evaluation responses using two independent judges, GPT-4o and DeepSeek-R1. Table~\ref{tab:judge-validation} reports the number of responses judged correct under each condition.

\begin{table}[t]
\centering
\small
\begin{tabular}{lccc}
\toprule
\textbf{Judge} & \textbf{No-Hint} & \textbf{Hier.-Hint} & \textbf{Oracle Hier.-Hint} \\
\midrule
o3-mini & $9/46$ & $16/46$ & $19/46$ \\
GPT-4o & $8/46$ & $14/46$ & $18/46$ \\
DeepSeek-R1 & $7/46$ & $13/46$ & $17/46$ \\
\bottomrule
\end{tabular}
\caption{Cross-judge evaluation of the same 138 responses across the No-Hint, Hier.-Hint, and Oracle Hier.-Hint conditions. The independent judges preserve the same ordering across conditions, while differing slightly in absolute strictness.}
\label{tab:judge-validation}
\end{table}

The paired treatment contrasts are stable across judges. Under o3-mini, Hier.-Hint improves accuracy over No-Hint by $15.22$ percentage points ($p=0.006$), while GPT-4o and DeepSeek-R1 both give a $13.04$ percentage-point improvement ($p=0.019$ and $p=0.009$, respectively). For Oracle Hier.-Hint, the improvement is $21.74$ percentage points under all three judges ($p<0.001$). Thus, although the judges differ slightly in absolute strictness, their relative treatment contrasts remain consistent.

Moving from o3-mini to GPT-4o to DeepSeek-R1, the No-Hint and Oracle Hier.-Hint counts each decrease by one problem per step, while Hier.-Hint decreases by two and then one. This largely uniform shift indicates that the judges differ mainly in overall strictness rather than in preference for a particular training condition. The one additional problem lost under Hier.-Hint thus bounds any potential judge-exploitation effect at a single problem, which leaves the significance of the treatment effect unchanged.

\paragraph{Human Validation.}
To further assess whether the judge threshold aligns with careful human evaluation, two authors with graduate-level mathematical training, blind to condition, seed, and judge score, jointly annotated a stratified sample of 25 responses in randomized order. The sample contained 20 randomly selected responses, which provide an unbiased estimate of agreement on the full evaluation set, and the 5 responses on which o3-mini and GPT-4o disagreed.

\begin{table}[t]
\centering
\footnotesize
\setlength{\tabcolsep}{3pt}
\begin{tabular}{lcc}
\toprule
\textbf{Stratum} &
$\mathbf{n}$ &
\shortstack{\textbf{o3-mini agrees}\\\textbf{with human}} \\
\midrule
Random (unbiased)
& $20$
& $19/20 = 95.0\%$ \\

o3-mini/GPT-4o disagree
& $5$
& $4/5 = 80.0\%$ \\
\midrule
Pooled
& $25$
& $23/25 = 92.0\%$ \\
\bottomrule
\end{tabular}

\caption{Human audit of the automated correctness judgments.
The random stratum provides the unbiased estimate of agreement,
while the disagreement stratum is deliberately over-sampled to
stress-test the judge.}
\label{tab:human-judge-validation}
\end{table}

Agreement on the unbiased random stratum is $95.0\%$ (Wilson 95\% CI $[76.4,99.1]$), and the pooled agreement is $92.0\%$ with Cohen's $\kappa=0.80$. Since the five judge-disagreement cases are deliberately over-sampled, the pooled figure serves as a conservative lower estimate. The two observed disagreements were split evenly, with one false positive and one false negative, indicating no systematic tendency of the judge to over- or under-credit responses at this sample size. As the responses were adjudicated jointly by both annotators, agreement is reported between the consensus label and the automated judge.

\section{Inference and Training Cost.}
The additional computational cost of H$^2$R primarily comes from one-time hint generation and LLM-based reward evaluation. For the 105 HiMath training problems, we use 105 o3-mini calls to generate the hierarchical hints, which are then reused throughout training. During training, approximately 4,200 reward-judge calls are issued per run, and approximately 1,334 additional judge calls are used for evaluation. Overall, the training and evaluation procedure requires approximately 5,700 LLM calls per run, in addition to the one-time 105 calls for hint generation.

\section{Inference-Only Results}
\label{app:inference-only-full}
 
\begin{center}
\includegraphics[width=\linewidth,
    height=0.22\textheight,
    keepaspectratio]{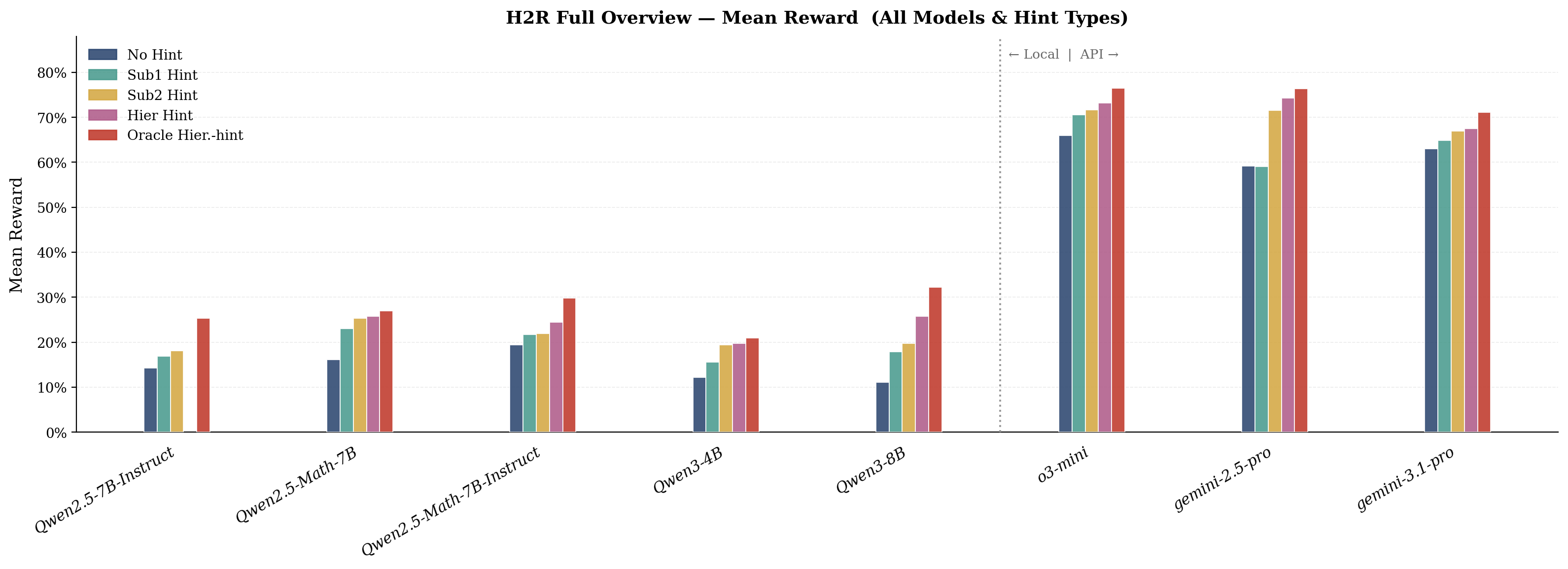}
\captionof{figure}{Inference-only mean reward across eight models
under five hint conditions without RL training.}
\label{fig:baseline_reward}
\end{center}
 
\begin{center}
\includegraphics[width=\linewidth,
    height=0.22\textheight,
    keepaspectratio]{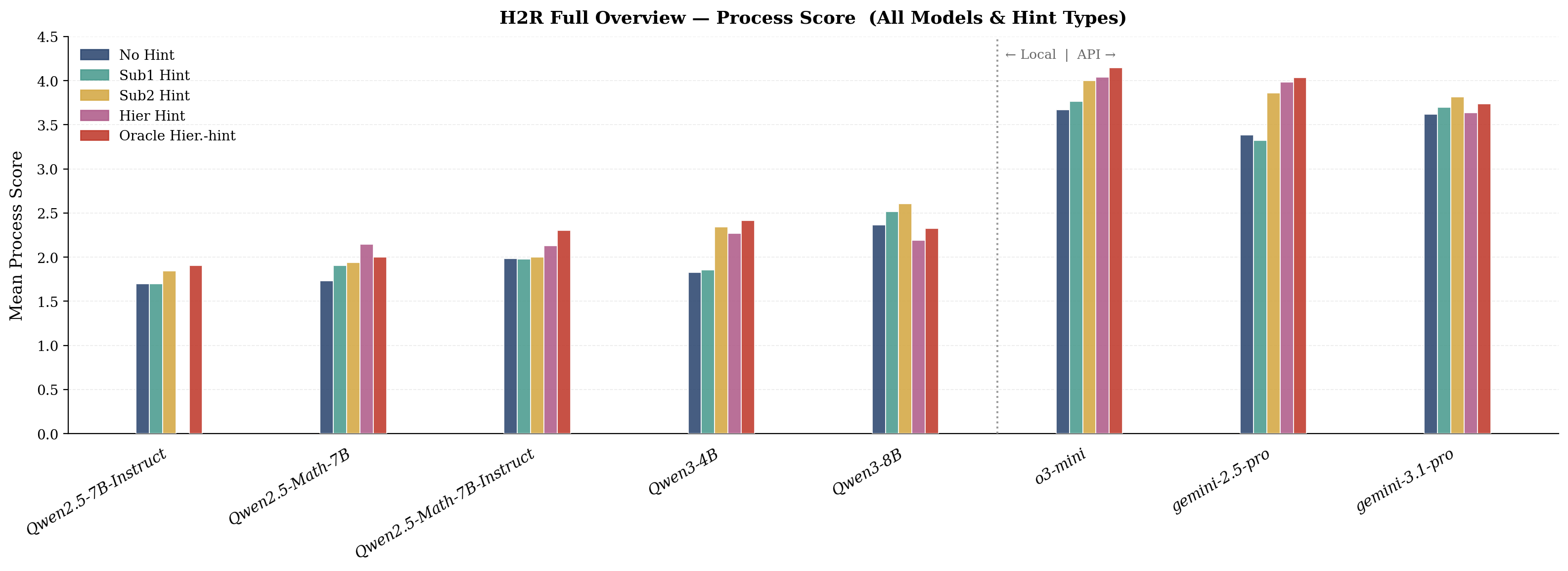}
\captionof{figure}{Inference-only mean process score across eight
models under five hint conditions without RL training.}
\label{fig:baseline_process}
\end{center}

\section{External-Benchmark Reasoning Analysis}
\label{app:external-reasoning}
We further analyze how H$^2$R changes the structure of model
reasoning on AIME24 and AIME25, which are outside the HiMath
training distribution. Both policies are evaluated without hints
at inference time.

\begin{table}[t]
\centering
\small
\setlength{\tabcolsep}{3pt}
\begin{tabular}{lccc}
\toprule
\textbf{Metric} &
\textbf{No-Hint} &
\textbf{H$^2$R} &
\textbf{$\Delta$ (95\% CI)} \\
\midrule
Length (tokens)
& $598 \pm 87$
& $1089 \pm 203$
& $+491$ $[413,570]$ \\

Step markers
& $1.3 \pm 0.8$
& $3.1 \pm 1.2$
& $+1.8$ $[1.3,2.3]$ \\

Interm. conclusions
& $2.8 \pm 1.4$
& $5.7 \pm 2.1$
& $+2.9$ $[2.0,3.8]$ \\

Sub-goal phrases
& $0.4 \pm 0.5$
& $1.8 \pm 0.9$
& $+1.4$ $[1.0,1.8]$ \\
\bottomrule
\end{tabular}

\caption{Lexical and structural characteristics of
Qwen2.5-7B-Instruct responses on AIME24 and AIME25
(60 problems, avg@5; 300 responses per condition).
No-Hint denotes the GRPO baseline, while H$^2$R denotes the
Hier.-Hint-trained policy. Both are evaluated without hints at
inference time. Confidence intervals are computed using paired
cluster bootstrap resampling with problems as clusters; all four
contrasts have $p<0.001$.}
\label{tab:reasoning-analysis}
\end{table}

On a representative AIME 2025 problem, the No-Hint policy asserts
two closed-form rules ``by standard circle properties'' without
deriving them and reaches an incorrect answer of $276^\circ$.
In contrast, the H$^2$R-trained policy derives the intermediate
steps, verifies that its six sub-results sum to $360^\circ$, and
reaches the correct answer of $336^\circ$. This difference suggests
that the change is not merely increased verbosity: the trained policy
produces more explicit intermediate states that can be checked during
reasoning. This observation is consistent with the error-propagation
analysis in Appendix~\ref{app:error-propagation}, where unreliable
intermediate states substantially reduce final-answer accuracy.

\section{Error Propagation in Hier.-Hint}
\label{app:error-propagation}
To quantify this propagation effect, we analyze
Qwen2.5-7B-Instruct under the Hier.-Hint condition by partitioning
problems into four groups based on whether the model's self-generated
$Q_1$ and $Q_2$ answers are correct, and reporting $Q_3$ accuracy
for each group. We focus on this model as it provides the most balanced
group sizes across all four conditions; stronger models rarely produce
incorrect intermediate answers, making propagation analysis statistically
underpowered for those settings.

\begin{figure}[t]
    \centering
    \includegraphics[width=\linewidth]{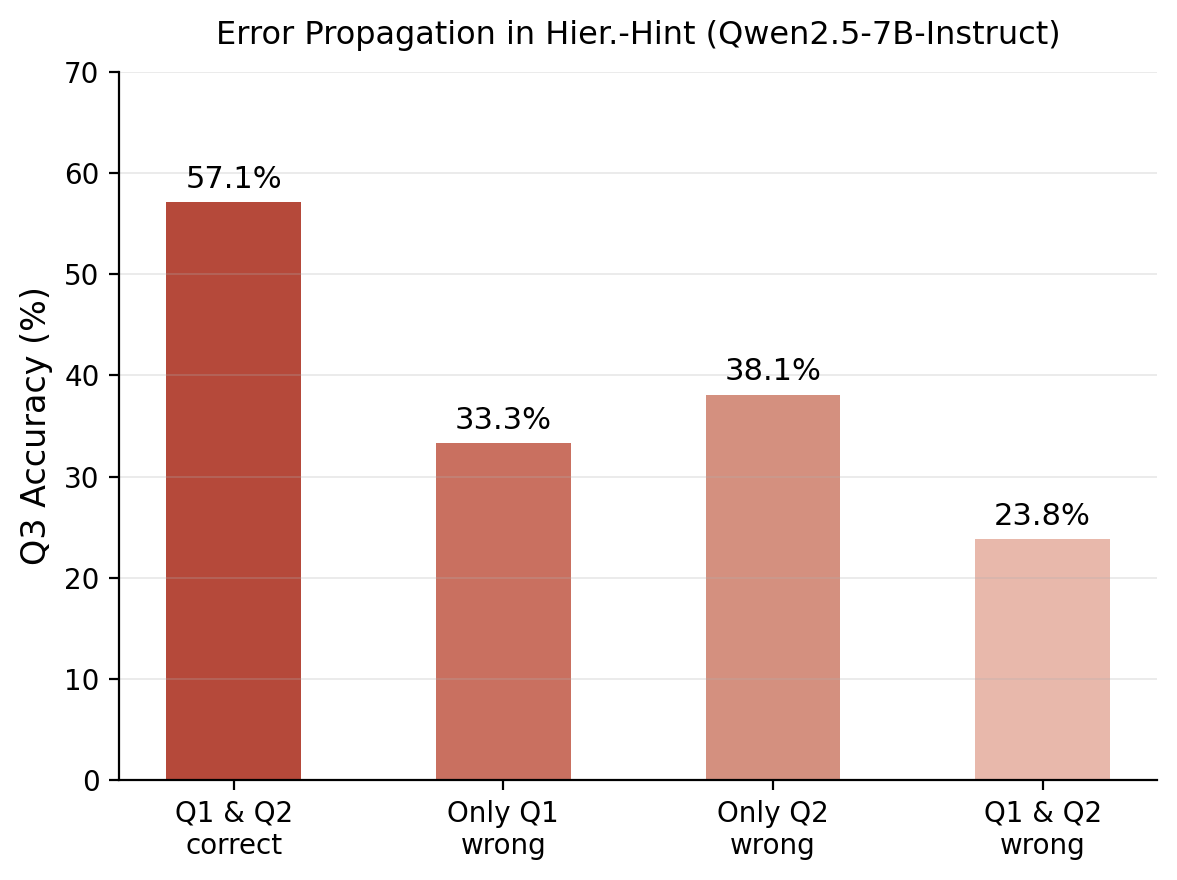}
    \caption{$Q_3$ accuracy under Hier.-Hint, dropping from 
    $57.1\%$ (both correct) to $23.8\%$ (both wrong).}
    \label{fig:error_propagation}
\end{figure}

In Figure~\ref{fig:error_propagation}, when both self-generated hints
are correct, $Q_3$ accuracy reaches 57.1\%. However, this drops to
33.3\% when only $Q_1$ is incorrect, and to 38.1\% when only $Q_2$
is incorrect. It declines further to 23.8\% when both hints are wrong.

These results yield three distinct observations. Crucially, error propagation is empirically confirmed and measurable, as the 33\% accuracy gap between the hint-correct and hint-incorrect conditions provides direct evidence that intermediate reasoning errors systematically undermine final-answer quality. Notably, $Q_1$ errors propagate more severely than $Q_2$ errors, causing a 24\% drop compared to a 19\% drop. This discrepancy indicates that mistakes in foundational reasoning steps are far more damaging than errors in intermediate conclusions. This pattern also aligns perfectly with the hierarchical structure of HiMath problems, where $Q_1$ establishes the definitional basis that all subsequent reasoning depends upon. 
Thirdly, even when both self-generated hints are incorrect, $Q_3$ retains a non-trivial accuracy of 23.8\%, indicating that the model retains some capacity for independent reasoning rather than blindly following its own prior outputs. Together, these findings suggest that closing the Hier.-Hint to Oracle Hier.-Hint gap requires improving the model's ability to generate reliable intermediate reasoning states. We denote this as a direction for future work on trainable hint generation between automated and oracle supervision.

\section{Attention Analysis}
\label{app:attention}
 
\vspace{4pt}
\begin{center}
\includegraphics[width=\linewidth,
    height=0.22\textheight,
    keepaspectratio]{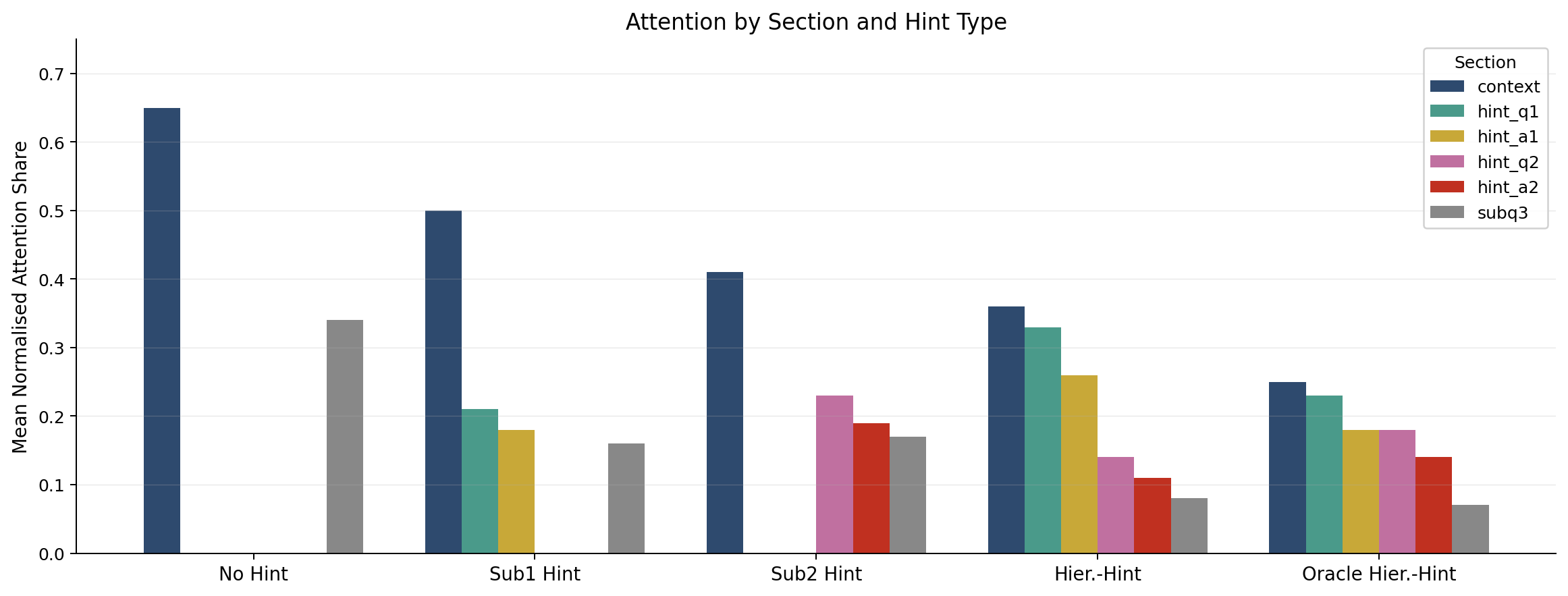}
\captionof{figure}{Mean normalised attention share across prompt
sections under five hint conditions.}
\label{fig:attention}
\end{center}
\vspace{2pt}
 
Three consistent patterns emerge. First, context attention decreases monotonically as hint informativeness increases (0.65 → 0.25), indicating that the model progressively shifts reliance from the raw problem statement to the provided intermediate guidance. Second, hint sections are attended to selectively and condition-specifically: Sub1 Hint activates only hint\_$Q_1$ and hint\_a1, Sub2-Hint activates only hint\_$Q_2$ and hint\_a2, and Hier./Oracle Hierarchical Hint engage all four hint sections simultaneously, confirming that the model correctly identifies relevant hint content rather than attending uniformly to all input tokens. Third, sub$Q_3$ attention drops substantially under richer hint conditions (0.34 → 0.07), suggesting that a more complete intermediate context reduces the need for repeated re-reading of the target question.  Fourth, under Hier.-Hint, the model allocates notably higher attention to hint\_$Q_1$ and hint\_a1 (0.33 and 0.26 respectively) compared to Oracle Hierarchical Hints (0.23 and 0.18), despite both conditions providing solutions to both subquestions. Together, these patterns provide mechanistic evidence that H$^2$R hints are actively integrated into the model's reasoning process rather than being ignored.
\section{Case Study}

\vspace{4pt}
\begin{center}
\includegraphics[width=\linewidth,
    height=0.18\textheight,
    keepaspectratio]{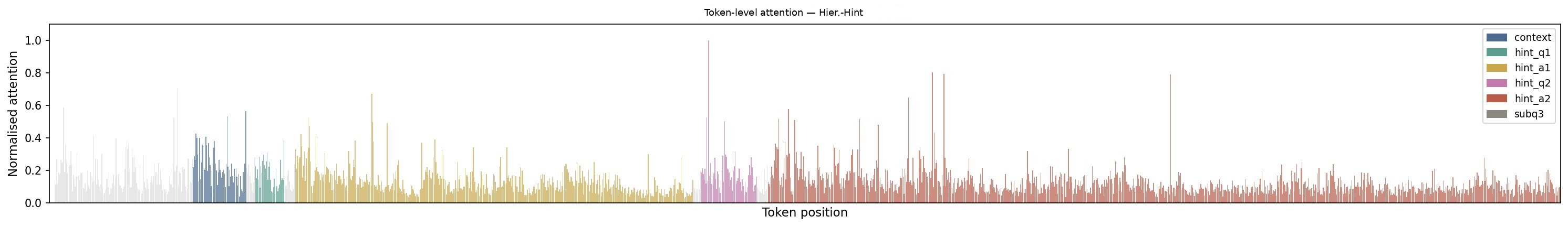}
\end{center}
\begin{center}
\includegraphics[width=\linewidth,
    height=0.18\textheight,
    keepaspectratio]{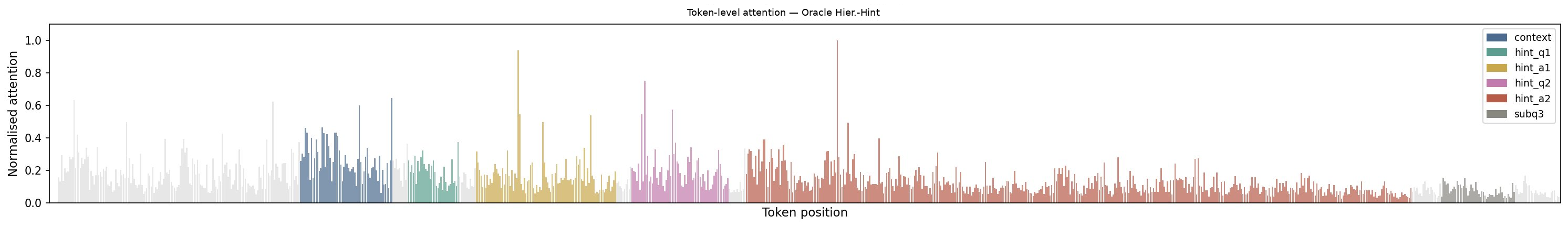}
\captionof{figure}{Token-level normalised attention under
Hier.-Hint (top) and Oracle Hier.-Hint (bottom). Colors: context (blue),
hint\_$Q_1$ (teal), hint\_a1 (yellow), hint\_$Q_2$ (pink),
hint\_a2 (red), sub$Q_3$ (dark grey). Under Hier.-Hint with
incorrect hints, attention peaks on the question text of $Q_2$
rather than its predicted answer, indicating failed hint
utilization. Under Oracle Hier.-Hint, sharp attention peaks emerge at
specific oracle answer tokens.}
\label{fig:attention_maps}
\end{center}
\paragraph{Figure construction.} Each bar in the figure corresponds to a single input token in the prompt. The horizontal axis represents token position, with tokens colour-coded by their prompt section: context (blue), hint\_$Q_1$ (teal), hint\_a1 (yellow), hint\_$Q_2$ (pink), hint\_a2 (red), and sub$Q_3$ (dark grey). The vertical axis reports the normalised attention value for each token, computed as follows: we extract attention weights from the final eight transformer layers during $Q_3$ generation, average across all attention heads and all generated output tokens, and apply min-max normalisation so that the most-attended token receives a value of 1.0. This produces a per-token measure of how much relative attention the model allocates to each part of the input while generating its answer.
\paragraph{Hier.-Hint (incorrect).} The most prominent attention peak (1.0) occurs within the hint\_$Q_2$ section — the text of subquestion 2 itself — rather than within hint\_a2, the model's self-generated answer to $Q_2$. The hint\_a1 section shows moderate, uniformly distributed attention without a sharp peak, and hint\_a2, despite occupying a large portion of the prompt, receives diffuse and low attention throughout. This pattern suggests that the model's self-generated $Q_2$ answer is of insufficient quality: unable to extract a reliable reasoning anchor from hint\_a2, the model reverts to attending to the question text of $Q_2$ instead. The absence of a focused attention peak within either answer section indicates that no clear intermediate reasoning step could be identified, and the model produces an incorrect final answer.
\paragraph{Oracle Hierarchical Hint (correct). } The attention pattern shifts markedly. The highest peak (1.0) now falls within hint\_a1, the ground-truth solution to $Q_1$, indicating that the model successfully identifies a specific, high-salience token within the oracle answer. The hint\_$Q_2$ section, which dominated under Hier.-Hint, now shows a substantially lower peak, suggesting that the model no longer needs to retreat to the question text when a reliable answer is available. Although hint\_a2 remains long, it exhibits several localised peaks of moderate height, indicating that the model extracts multiple relevant reasoning cues from the correct $Q_2$ solution rather than scanning it diffusely.

\paragraph{Interpretation.} The contrast between the two conditions reveals that the performance gap between Hier.-Hint and Oracle Hierarchical Hint is not a consequence of different attention strategies, but of different hint quality. In both conditions, the model actively attends to hint content. However, when self-generated hints are incorrect, the model cannot form focused attention peaks within the answer sections and is forced to fall back on the question text — a behavioural signature of failed hint utilisation. When oracle hints are provided, sharp attention peaks emerge at specific answer tokens, providing the reasoning anchors necessary for a correct final answer. This finding directly supports the claim that hint quality, rather than the presence of hints per se, is the primary determinant of Hier.-Hint performance.

\end{document}